\documentclass[runningheads]{llncs}
\usepackage[T1]{fontenc}
\usepackage{graphicx}
\usepackage{tabularx}
\usepackage{makecell}
\usepackage{multirow}
\usepackage{enumitem}
\usepackage{amsmath}
\usepackage{orcidlink}
\begin{document}
\title{Confidence-Aware and Self-Supervised \\Image Anomaly Localisation}
\titlerunning{Confidence-Aware Image Anomaly Detection}
%
\author{Johanna P. M\"uller \inst{1}\orcidlink{0000-0001-8636-7986} \and
Matthew Baugh\inst{2}\orcidlink{0000-0001-6252-7658} \and
Jeremy Tan\inst{3}\orcidlink{0000-0002-9769-068X} \and
Mischa Dombrowski\inst{1}\orcidlink{0000-0003-1061-8990} \and
Bernhard Kainz\inst{1,2}\orcidlink{0000-0002-7813-5023}}
\authorrunning{ M\"uller et al.}
%
\institute{
Friedrich–Alexander University Erlangen–N\"urnberg, DE 
\email{johanna.paula.mueller@fau.de} \and
Imperial College London, SW7 2AZ, London, UK \and
ETH Zurich, CH}

%
\maketitle              
\begin{abstract}
Universal anomaly detection still remains a challenging problem in machine learning and medical image analysis. It is possible to learn an expected distribution from a single class of \emph{normative samples}, \emph{e.g.}, through epistemic uncertainty estimates, auto-encoding models, or from synthetic anomalies in a self-supervised way. 
The performance of self-supervised anomaly detection approaches is still inferior compared to methods that use examples from \emph{known unknown} classes to shape the decision boundary. However, outlier exposure methods often do not identify \emph{unknown unknowns}.  
Here we discuss an improved self-supervised single-class training strategy that supports the approximation of probabilistic inference with loosen feature locality constraints. We show that up-scaling of gradients with histogram-equalised images is beneficial for recently proposed self-supervision tasks. 
Our method is integrated into several out-of-distribution (OOD) detection models and we show evidence that our method outperforms the state-of-the-art  on various benchmark datasets.

\end{abstract}

\section{Introduction}

Out-of-distribution (OOD) detection builds upon the assumption that the division into normal and abnormal data is distinct, however, OOD data can overlap in-distribution (ID) data and may exhibit an infinite number of descriptive features. 
We assume for medical imaging data a finite ID ("healthy") distribution space and an infinite OOD ("anomalous") distribution space. Furthermore, we assume ID consistency for healthy medical images such that the compatibility condition holds, based on the impossibility theorems for OOD detection by~\cite{fang2022out}. As a result, OOD detection algorithms can be capable of learning the finite ID space and also a finite but sufficient number of ODD features for inference. We can approximate density-based spaces based on drawn samples from real unknown (conditioned) probability distributions for covering uncertainty in the annotation of data, and, therefore, assume the Realisability assumption~\cite{fang2022out} for learnable OOD detection referring to the proposed problem formulation.

The OOD problem for medical imaging can also be seen from a practical, intuitive point of view. To reflect that multiple human medical experts can come by different diagnoses given the same image of a patient, we integrate uncertainty estimates for both ID and OOD data in the form of probability distributions.
Intuitively, we tend to imagine a finite ID space, since we observe a consistency between ID features which are exhibited by healthy human individuals from an anatomical point of view. By assuming that, we postulate that we can present learnable OOD detection through training different types of algorithms on normal data with synthetically generated anomalies. 

Learning from synthetically generated anomalies became a research focus in medical image analysis research recently~\cite{han2019synthesizing}.  In a medical context, labelling requires medical expertise and, hence, human resources for generating reliable ground truth masks for anomaly detection algorithms. Self-supervised tasks that base on synthetically generated anomalies are considered convenient mitigation for  limited robustness and generalisation abilities  that result from small datasets. 
An extension of this idea is to leverage the natural variations in normal anatomy to create a range of synthetic abnormalities. For example, image patch regions can be extracted from two independent samples and replaced with an interpolation between both patches~\cite{tan2020detecting,li2021cutpaste}. The interpolation factor, patch size, and patch location can be randomly sampled from uniform distributions. Any encoder-decoder architecture can be trained to give a pixel-wise prediction of the patch and its interpolation factor. This encourages a deep network to learn what features to expect normally and to identify where foreign patterns have been introduced. The estimate of the interpolation factor lends itself nicely to the derivation of an outlier score. Meanwhile, the pixel-wise output allows for pixel- and subject-level predictions using the same model. However, such synthesis strategies feature obvious discontinuities. 
\cite{tan2021detecting,schluter2021self} solve the discontinuity problem by using Poisson image editing, but the resulting anomalies can be so subtle that they may represent variations of the normal class rather than true anomalies and these approaches do not provide prediction confidence estimates. 
Therefore we propose a new approach to model the ID space and make the following contributions:
\begin{enumerate}[noitemsep]
\item  We propose a revised Poisson Image-interpolation framework for the generation of salient but still smoothly interpolated anomalies for self-supervision in unsupervised image anomaly localisation.
\item We propose self-supervision with a probabilistic feature extractor -- Probabilistic PII (P-PII) -- which allows the generation of stochastic  anomalies with which we are able to simulate multiple annotators.
\item We evaluate P-PII on 2D chest radiograph images and 3D CT scans and show that our method outperforms recently proposed self-supervised anomaly localisation approaches.  
\item We show that it is possible to learn feature distributions for 'normal' tissue in a self-supervised way from databases that exclusively contain patients with the disease.
\end{enumerate}

\noindent\textbf{Related Work.}
\sloppy
The most prominent direction for unsupervised medical anomaly localisation~\cite{tschuchnig2022anomaly} is dominated by reconstruction-based methods like VAEs~\cite{venkatakrishnan2020self,zhou2020unsupervised,li2020out,guo2021cvad} 
as well as other generative models like GANs~\cite{zenati2018efficient,akcay2018ganomaly,schlegl2019f}
, especially, for image synthesis and data augmentation~\cite{chen2018deep,han2019synthesizing,guan2019breast}. New advances are expected by Diffusion models, which shine with detailed reconstructions and anomaly maps for detection~\cite{wolleb2022diffusion} but they are computationally very challenging and have not been evaluated in detail yet. Other commonly used methods include one-class Support Vector Machines, k-Nearest Neighbors and extensions of these approaches for dimensionality-reduced feature spaces~\cite{liang2017enhancing,cao2020benchmark}.
Probabilistic methods have not been researched in detail for OOD detection yet. However, they are known for probabilistic segmentation approaches. For example, the Probabilistic Hierarchical Segmentation (PHISeg) combines a conditional variational autoencoder (cVAE) with a U-NET setup proposed by \cite{baumgartner2019phiseg}, Bayesian U-Nets~\cite{seebock2019exploiting} can model epistemic uncertainty with weak labels and Monte Carlo estimates~\cite{pawlowski2018unsupervised,baur2020bayesian,nakao2022anomaly}.

In a medical context, labelling requires medical expertise and, hence, human resources for generating reliable ground truth masks for anomaly detection algorithms. Self-supervised tasks are considered as convenient extensions for improving robustness, uncertainty and generalisation abilities of models and replace expensive labelling~\cite{hendrycks2019using,henaff2020data,mohseni2020self,zhao2021anomaly}. We modify our backbone models to allow for OOD detection. To do this, we form a self-supervised task which is easily interchangeable. The self-supervised principle relies on patch interpolation from the same or a different source image into a target image. Since more research work focuses on alleviating the labelling effort by experts for image data, different generation methods for anomalies emerged.
For Foreign patch interpolation (FPI)~\cite{tan2020detecting}, two patches of the same location are extracted from
two independent samples and replaced with an interpolation between both patches. CutPaste~\cite{li2021cutpaste} updates the original method by translating patches within an image and allows the effective detection of anomalies in industrial datasets.
Poisson Image Interpolation (PII)~\cite{tan2021detecting} overcomes sharp discontinuities with Poisson editing as an interpolation strategy and generates more organic and subtle outliers. Natural Synthetic Anomalies (NSA)~\cite{schluter2021self} are introduced by rescaling, shifting and a new Gamma-distribution-based patch shape sampling without the use of interpolation factors for an end-to-end model for anomaly detection.

\section{Method}

\begin{figure}[t]
\centering
\includegraphics[width=\textwidth]{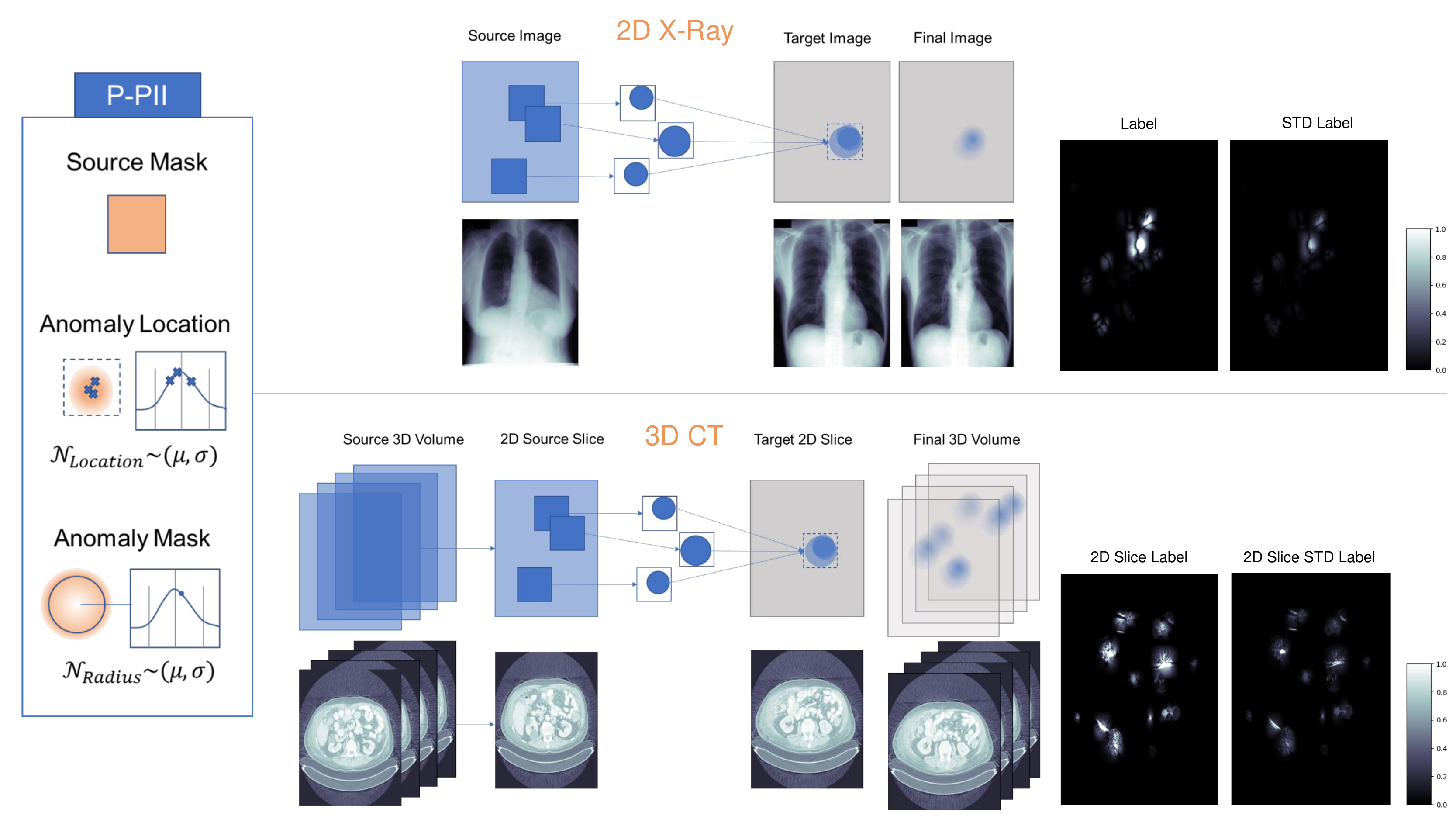}
\caption{Probabilistic PII takes patches from a source image of a given size. A second mask of circular size, drawn from two normal distributions for radius and location inside the source patches, allows aggregated anomalies with smoothly interpolated boundaries. We obtain probabilistic and salient anomalies.}
\label{fig:scheme}
\end{figure}

Self-supervised tasks were considered convenient extensions for improving the robustness, uncertainty and generalisation abilities of models \cite{hendrycks2019using,henaff2020data,mohseni2020self}. Our proposed Probabilistic PII self-supervision task is based on~\cite{tan2020detecting} and builds upon the Poisson image editing implementation by~\cite{Baugh_PIE-torch}. 
PII relies on the relative changes of the source image, the image gradient field $\mathbf{v_{pq}}$, in the patch region and the patch boundary of the target image $\delta h$, see Eq. \ref{v}. The solution of the underlying mathematical problem represents the discretised Poisson equation with Dirichlet boundary conditions, see Eq. \ref{nabla} and Eq. \ref{direchlet}.
The intensity values within the patch $h$ are given by the scalar function $f_{in}$ and $\langle p,q \rangle$ are denoted as a pixel pair such that $q \in N_p$ denote the four directly adjacent neighbour pixel of $p$. For PII, $\alpha$ determines the influence of the individual image gradients on the interpolation task.

\begin{equation}\label{v}
v_{pq} =\left\{ 
    \begin{array}{ c l }
        (1 - \alpha) (x_{i_{p}} - x_{i_{q}}), & {\quad \textrm{if } \left|(1 - \alpha) (x_{i_{p}} - x_{i_{q}})\right| > \left|\alpha (x_{i_{p}} - x_{i_{q}})\right|}\\
        \alpha (x_{i_{p}} - x_{i_{q}}),     & \quad \textrm{otherwise.}
    \end{array}
\right.
\end{equation}

\begin{equation}\label{nabla}
\mathcal{r} f_{in} = \textrm{div} \mathbf{v} \textrm{ over } h
\end{equation}

The PII task can be reformulated to the following minimisation problem (Eq.~\ref{min}), given the projection of $\mathbf{v}(\frac{p+q}{2})$ onto the oriented edge (Eq.~\ref{v}) \cite{tan2021detecting} and the field of intensity image gradients (Eq.~\ref{nabla}). The problem formulation can be solved via a discrete linear system solver. 

\begin{equation}\label{min}
\min_{f_{in}} \iint\displaylimits_h \left| {\mathcal{r} f_{in} - \mathbf{v}} \right| ^2, 
\textrm{  with } f_{in} \Big|_{ \delta h} = f_{out} \Big|_{ \delta h} 
\end{equation}

\begin{equation}\label{direchlet}
\min_{f_{in}\big|_{h}} \sum_{\langle p,q \rangle \cap h \neq 0} ( f_{in,p} - f_{in,q} - v_{pq})^{2},
\textrm{  with } f_{in} \Big|_{ \delta h} = f_{out} \Big|_{ \delta h}, \forall p \in \delta h, q \in N_p
\end{equation}

Our proposed Probabilistic PII (P-PII) builds upon these mathematical foundations but incorporates new features and approaches for addressing current limitations and rethinking its application.

First, we apply P-PII pairwise on allegedly non-anomalous training data samples but those pairs can be also, \emph{e.g.}, easily reduced to one and single non-anomalous image sample for reduced memory and time consumption. If pairwise applied, the allocation of image pairs is randomly drawn from the image batch. Second, we take patches from different locations of the source image and interpolate them into different locations inside the target image, hence, we latch on the patch drawing by NSA~\cite{schluter2021self}. Third, we overcome the current limitation of PII and PII-based anomaly generation methods regarding the grade of abnormality of the interpolated patches. If both, source and target images, are normalised, these anomalous regions are very subtle and difficult to recognise - compared to real lesions as well. For intensifying these abnormal features, we introduce an amplification of gradients, through a scaling factor, during the interpolation into the source patch. This approach generates less subtle, salient anomalies which are still smoothly interpolated into the target image. Fourth, we mitigate class imbalance of normal and anomalous pixels through the generation of $k > 1$ anomalies per image with which we speed up learning to differentiate both classes. Fifth, we introduce the Probabilistic feature into PII.
For simulating the variance of annotations by multiple raters, as \emph{e.g.}, annotation of lesions by multiple medical experts, we generate circular anomalies, inside each extracted patch from the source image. Therefore, we draw anomaly masks whose parameters, radius $\textbf{r}$ and location $\textbf{(x,y)}$ (Eq. \ref{norm}), we sample from normal distributions. We ensure with fixed boundaries of location and radius that the generated anomaly only touches the boundaries. 

\begin{equation}\label{norm}
\textbf{r} \sim \mathcal{N}_{Radius}(\mu,\sigma) \quad \quad
(\textbf{x},\textbf{y}) \sim \mathcal{N}_{Location}(M=\langle\mu_\textbf{x},\mu_\textbf{y}\rangle,\Sigma=\langle\sigma_\textbf{x},\sigma_\textbf{y}\rangle)
\end{equation}

For using P-PII as the self-supervised task for OOD detection, we decided on intensity-based label generation. Based on the mean of all anomalies of each patch, we use the absolute difference between the original target image and the mean final image as the label. Additionally, we have a variance map of all anomalies which can be used for further statistical evaluation or integration into the optimisation problem. 

\section{Evaluation and Results}

\noindent\textbf{Data.} 
We use the \href{http://db.jsrt.or.jp/eng.php}{JSRT} database \cite{shiraishi2000development} as an exemplary smaller medical imaging dataset which includes 154 conventional chest radiographs with lung nodules and 93 radiographs without a nodule. For each patient, only one image is attributed. We re-scaled all images from $2048\times2048$ matrix size to $512\times512$ in order to hold the conditions for all datasets equal. The subset without pathological findings serves as our training dataset. 
\href{https://wiki.cancerimagingarchive.net/pages/viewpage.action?pageId=1966254}
{LIDC-IDRI}~\cite{armato2011lung} covers 1018 cases in the form of CT scans with 7371 lesions, which were marked by at least one radiologist. We also divide the dataset into lesion slices and anomaly-free slices by extracting the context slices from each volume with a margin of about 5 slices on either side of the lesion, which approximates the maximum possible margin of lesions given slice thickness and lesion diameter. We use the first 800 cases as a training dataset, and the rest for validation and testing.
The large-scale dataset \href{https://nihcc.app.box.com/v/DeepLesion}
{DeepLesion}~\cite{yan2018deeplesion} contains 32,735 lesions in 32,120 computed tomography (CT) slices from 10,594 studies of 4,427 unique patients. Since the image size varies throughout the dataset, we resize each image to the smallest occurring size, $512\times512$. Each lesion slice is provided as part of an imaging volume which provides the 3D context of the lesion. We divide the dataset into lesion slices and anomaly-free slices by extracting the context slices from each volume with a margin of about 10 mm on either side of the lesion. As a result, we have 386,587 anomaly-free slices and 4831 annotated anomalous slices. We test the quality of performance for all models on ID and OOD data samples, which were not seen during training. For JSRT, the test set consists of $19$ ID samples and $154$ OOD samples. For the large datasets, we drew a test cohort of $500$ ID and $500$ ($478$ for LIDC-IDRI) OOD samples. For LIDC-IDRI and DeepLesion, both ID and OOD samples are from patients not occurring in the training dataset. 
Note that the models are trained on healthy tissue of ill patients for the datasets LIDC-IDRI and DeepLesion, which is different to the dataset JSRT for which we only differentiate between ill and healthy patients/samples. 

\noindent\textbf{Pre-processing and Training.}
We use histogram equalisation to the normalised images for contrast enhancement, adopted from MIMIC-CXR-JPG \cite{johnsonmimic}. We apply this type of equalisation to all datasets. We train all models for a fixed number of $100,000$ steps with a fixed batch size of $16$. We used PNY NVIDIA A100s with at least 18 GB of memory per job. The training runtime was approx. 4 days. The backbone models and P-PII were implemented in Python and TensorFlow.

\noindent\textbf{Metrics.} 
Choosing suitable metrics for evaluating OOD detection methods is important to effectively evaluate the performance of a method and make valid comparisons with other approaches. 
 We chose the Area under the receiver operating characteristic (AUROC) for sample- and pixel-wise binary classification between OOD and ID samples/pixels as a threshold-less metric. We refer with \emph{OOD} to anomalous samples/pixels and with \emph{ID} to normal ('healthy') input samples/pixels.
 Average Precision (AP) takes both precision and recall into account and is considered a sample-based evaluation metric here.
In medical imaging analysis, false negatives are more critical than false positives, especially, in lesion detection. Therefore, we include the Free-response receiver operating characteristic (FROC) score as an evaluation measure. 

\noindent\textbf{Sensitivity analysis.}
We perform an ablation study to investigate the impact of revised PII as a self-supervision task for various backbone models (U-Net, Monte-Carlo Dropout (rate=0.1) U-Net, PHiSeg).
All backbone models have the same depth of five levels, PHiSeg includes two additional resolution levels. We examine the influence of selected augmentation functions for small-scale datasets or datasets suffering from class imbalance for improving the performance of self-supervised training.  

\begin{figure}[h!]
\centering
\includegraphics[width=\textwidth]{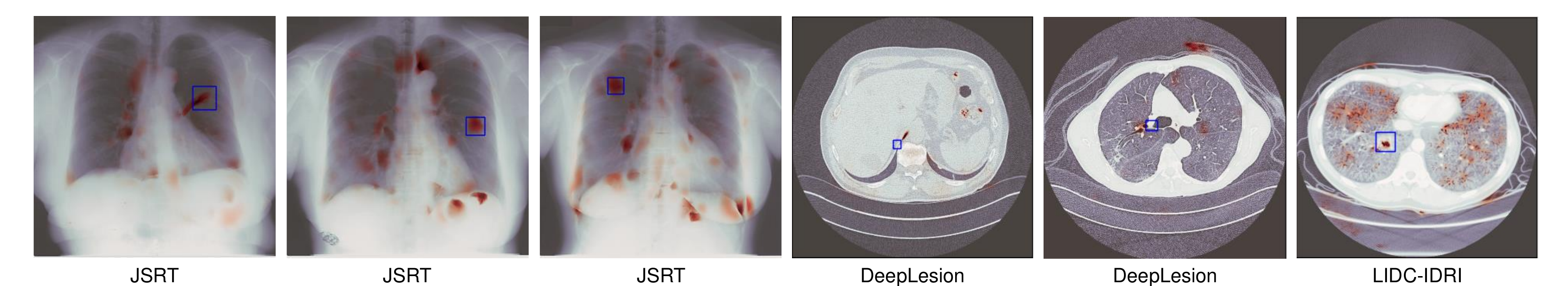}
\caption{Exemplary anomaly prediction on test data with U-net, input image in grey, heatmap of prediction in red, ground truth bounding box in blue.}
\label{fig:results}
\end{figure}

\noindent\textbf{Results.} 
We evaluated all models with the training checkpoint for best dice. We show quantitative results in Tab.~\ref{tab:quantitative} for all backbone models. We observed an increase of pixel-wise AUROC of up to $13 \%$ for U-net and PHiSeg and $18 \%$ for Dropout U-net, for the JSRT dataset. For LIDC-IDRI, we achieve values improved by up to $53 \%$ for PHiSeg. For DeepLesion, we determined an increase of $34 \%$ with PHiSeg and $9 \%$ with U-net for pixel-wise AUROC.
Emphasising the sensitivity level of $0.27$ for 10 avg. FPS, we increased the performance of the U-net, trained with PII, threefold with our proposed self-supervision task. 
Sample-wise AUROC was improved the most for the JSRT dataset with $45 \%$, whereas we observed AUROC values $< 0.5$ for LIDC-IDRI and, partially, for DeepLesion and JSRT. 
An increased amount of false positives in predicting anomalous samples results for sample-wise AP for the large datasets. 
We show qualitative results for the prediction of U-net as a backbone model in Fig.~\ref{fig:results}. The prediction on JSRT is quantitatively better, but there are still false positive pixels in all examples, especially, for the larger datasets.
We compare augmentation functions for further enhancing the performance of P-PII, see Tab.~\ref{tab:augmentation}. We compare both best-performing models and obtain an increase of $1 \%$ with scaling of the input image and combining scaling, random rotation in between $\pm10^{\circ}$ and elastic deformation. Further improvement was achieved by scaling the input for the Dropout U-net which resulted in enhancing image-wise AUROC about $3 \%$. The highest improvement can be achieved through the use of augmentation functions yielding a sensitivity of $11 \%$ for U-net with combined augmentation, and $19\%$ for Dropout U-net with scaling.

\begin{table}[t]
\caption{Results; for PHiSeg, mean of $50$ drawn samples from likelihood network; AUC - Area under the Receiver operating characteristic (AUROC), FC - Free-response Receiver operating characteristic (FROC) for $10$ average FPs}
\centering
\begin{tabularx}{\textwidth}{clccccccccccccc}
\toprule
&& & \multicolumn{4}{c}{JSRT~\cite{shiraishi2000development}} & \multicolumn{4}{c}{DeepLesion~\cite{yan2018deeplesion}} & \multicolumn{4}{c}{LIDC-IDRI~\cite{armato2011lung}}\\
&& & \multicolumn{2}{c}{Pixel} & \multicolumn{2}{c}{Sample} & \multicolumn{2}{c}{Pixel} & \multicolumn{2}{c}{Sample} & \multicolumn{2}{c}{Pixel} & \multicolumn{2}{c}{Sample}\\
\cmidrule(lr){4-7}\cmidrule(lr){8-11}\cmidrule(lr){12-15}
 \multirow{4}{*}{\rotatebox{90}{PII~\cite{tan2020detecting}}} &Model &  & AUC &  FC & AUC & AP & AUC & FC & AUC & AP & AUC &  FC & AUC & AP \\ 
&U-Net &  & $0.80$ & $0.08$ & $0.44$ & $0.87$ & $0.68$ & $0.00$ & $0.50$ & $0.49$ & $0.50$ & $0.00$ & $0.36$ & $0.39$ \\
&MC U-Net &  & $0.76$ & $0.01$ & $0.55$ & $0.90$ & $\mathbf{0.74}$ & $0.00$ & $0.53$ & $\mathbf{0.55}$ & $0.59$ & $\mathbf{0.01}$ & $0.40$ & $0.43$ \\
&PHiSeg &  & $0.67$ & $0.00$ & $0.51$ & $0.90$ & $0.41$ & $\mathbf{0.01}$ & $0.47$ & $0.48$ & $0.43$ & $0.00$ & $\mathbf{0.52}$ & $\mathbf{0.50}$ \\
\midrule
\multirow{3}{*}{\rotatebox{90}{\makecell[l]{Ours \\P-PII}}}&U-Net &  & $\mathbf{0.90}$ & $\mathbf{0.27}$ & $\mathbf{0.64}$ & $\mathbf{0.94}$ & $\mathbf{0.74}$ & $\mathbf{0.01}$ &  $\mathbf{0.56}$ & $0.52$  & $\mathbf{0.69}$ & $\mathbf{0.01}$ &  $0.33$ & $0.38$  \\
&MC U-Net & & $\mathbf{0.90}$ & $0.26$ & $\mathbf{0.64}$ & $0.93$ & $0.72$ & $\mathbf{0.01}$ &  $0.47$ & $0.49$ & $0.67$ & $\mathbf{0.01}$ &  $0.38$ & $0.41$ \\
&PHiSeg &  & $0.76$ & $0.06$ & $0.63$ & $0.93$ & $0.55$ & $\mathbf{0.01}$ &  $0.55$ & $0.51$  & $0.66$ & $\mathbf{0.01}$ &  $0.41$ & $0.44$  \\
\bottomrule
\end{tabularx} 
\label{tab:quantitative}
\vspace{0cm}
\end{table}

\begin{table}[h]
\caption{Sensitivity analysis of augmentation functions for small-scale datasets on P-PII for JSRT~\cite{shiraishi2000development}; scaling, combined (rotation $\pm10^{\circ}$, elastic deformation, scaling).}
\centering
\begin{tabularx}{\textwidth}{llcXccXcXc}
\toprule
 &&&&\multicolumn{2}{c}{AUROC}&&\multicolumn{1}{c}{AP}&&\multicolumn{1}{c}{FROC}\\
 & Model& Augmentation && Pixel & Image && Image & & 10FPs  \\
\cmidrule(lr){5-6}\cmidrule(lr){8-8}\cmidrule(lr){10-10}
\multirow{4}{*}{\rotatebox{90}{\makecell[l]{Ours \\P-PII}}} & U-Net & scaling && $\mathbf{0.91}$ & $0.60$ && $0.93$ &&$0.27$  \\
&MC U-Net & scaling && $\mathbf{0.91}$ & $\mathbf{0.66}$ && $\mathbf{0.94}$ && $\mathbf{0.31} $ \\
 \cmidrule(lr){2-10}\
&U-Net & combined && $\mathbf{0.91}$ & $0.60$ && $0.92$ && $0.30$  \\
&MC U-Net & combined && $\mathbf{0.91}$ & $0.59$ && $0.93$ & &$0.25$  \\
\bottomrule
\end{tabularx} 
\label{tab:augmentation}
\vspace{-0.5cm}
\end{table}

\noindent\textbf{Discussion.}
Self-supervision with P-PII enables all models to detect also very small lesions, see Fig.~\ref{fig:results}, which is still a major challenge for other anomaly localisation models, in both, a supervised and self-supervised context.
We improve upon the issue of decreasing sensitivity for increasing average FPs in FROC, which we observe for the baseline method. 
With augmentation functions the performance of models trained with PII increases the sensitivity significantly by up to $19\%$.
The limited quantitative performance on DeepLesion and LIDC-IDRI is likely due to the fixed training steps which could be insufficient for large datasets and also the foreground-background class imbalance could influence the results for large datasets. These issues need to be approached in further studies.
Considering the number of false positive predicted regions, we would require expert analysis if those regions are correlated with real aberrations in the input images. For now, we can only interpret them as visually perceived abnormal regions in the input images, \emph{e.g.}, dense regions in the lung hilum.
Compared to the original PII implementation we achieved a shortening of at least half of the training time through the usage of Poisson image editing through discrete sine transformation~\cite{Baugh_PIE-torch}. This allows us to sample from different source images multiple times for probabilistic representations of anomalies while still being faster than the baseline.

\section{Conclusion}

We analyse the proposed self-supervised learning method, P-PPI, on multiple three backbone models and three small- and large-scale datasets from the medical imaging domain. We exploit the influence of augmentation functions for the self-supervision task and present probabilistic anomalies, which are described for the first time for applications in OOD detection. 
Our investigations highlight previous limitations when using Poisson image interpolation for the generation of synthetic anomalies. 
We improve pixel-wise AUROC by up to $18 \%$ and sample-wise AUROC by up to $45 \%$ in comparison to baseline methods. Additionally, we enhanced the pixel-wise sensitivity to $10$ avg. FPs up to $38 \%$. We also show that it is possible to learn feature distributions for normal tissue in a self-supervised way from databases that exclusively contain patients with the disease (DeepLesion and LIDC-IDRI). 
In future work, the integration of the generated variance maps into the loss function has a high potential for pushing unsupervised probabilistic learning further towards integration into clinical workflows. \\ 

\noindent\emph{Acknowledgements}:  The authors gratefully acknowledge the scientific support and HPC resources provided by the Erlangen National High Performance Computing Center (NHR@FAU) of the Friedrich-Alexander-Universität Erlangen-Nürnberg (FAU) under the NHR projects b143dc and b180dc. NHR funding is provided by federal and Bavarian state authorities. NHR@FAU hardware is partially funded by the German Research Foundation (DFG) – 440719683. Additional support was also received by the ERC - project MIA-NORMAL 101083647,  DFG KA 5801/2-1, INST 90/1351-1 and by the state of Bavaria.
%
%
%
\bibliographystyle{splncs04}
\bibliography{paper}

\begin{thebibliography}{10}
\providecommand{\url}[1]{\texttt{#1}}
\providecommand{\urlprefix}{URL }
\providecommand{\doi}[1]{https://doi.org/#1}

\bibitem{akcay2018ganomaly}
Akcay, S., Atapour-Abarghouei, A., Breckon, T.P.: Ganomaly: Semi-supervised
  anomaly detection via adversarial training. In: Asian conference on computer
  vision. pp. 622--637. Springer (2018)

\bibitem{armato2011lung}
Armato~III, S.G., McLennan, G., Bidaut, L., McNitt-Gray, M.F., Meyer, C.R.,
  Reeves, A.P., Zhao, B., Aberle, D.R., Henschke, C.I., Hoffman, E.A., et~al.:
  The lung image database consortium (lidc) and image database resource
  initiative (idri): a completed reference database of lung nodules on ct
  scans. Medical physics  \textbf{38}(2),  915--931 (2011)

\bibitem{Baugh_PIE-torch}
Baugh, M.: {PIE-torch},
  \url{https://github.com/matt-baugh/pytorch-poisson-image-editing}

\bibitem{baumgartner2019phiseg}
Baumgartner, C.F., Tezcan, K.C., Chaitanya, K., H{\"o}tker, A.M., Muehlematter,
  U.J., Schawkat, K., Becker, A.S., Donati, O., Konukoglu, E.: Phiseg:
  Capturing uncertainty in medical image segmentation. In: International
  Conference on Medical Image Computing and Computer-Assisted Intervention. pp.
  119--127. Springer (2019)

\bibitem{baur2020bayesian}
Baur, C., Wiestler, B., Albarqouni, S., Navab, N.: Bayesian skip-autoencoders
  for unsupervised hyperintense anomaly detection in high resolution brain mri.
  In: 2020 IEEE 17th International Symposium on Biomedical Imaging (ISBI). pp.
  1905--1909. IEEE (2020)

\bibitem{cao2020benchmark}
Cao, T., Huang, C.W., Hui, D.Y.T., Cohen, J.P.: A benchmark of medical out of
  distribution detection. arXiv preprint arXiv:2007.04250  (2020)

\bibitem{chen2018deep}
Chen, X., Pawlowski, N., Rajchl, M., Glocker, B., Konukoglu, E.: Deep
  generative models in the real-world: An open challenge from medical imaging.
  arXiv preprint arXiv:1806.05452  (2018)

\bibitem{fang2022out}
Fang, Z., Li, Y., Lu, J., Dong, J., Han, B., Liu, F.: Is out-of-distribution
  detection learnable? arXiv preprint arXiv:2210.14707  (2022)

\bibitem{guan2019breast}
Guan, S., Loew, M.: Breast cancer detection using synthetic mammograms from
  generative adversarial networks in convolutional neural networks. Journal of
  Medical Imaging  \textbf{6}(3),  031411 (2019)

\bibitem{guo2021cvad}
Guo, X., Gichoya, J.W., Purkayastha, S., Banerjee, I.: Cvad: A generic medical
  anomaly detector based on cascade vae. arXiv preprint arXiv:2110.15811
  (2021)

\bibitem{han2019synthesizing}
Han, C., Kitamura, Y., Kudo, A., Ichinose, A., Rundo, L., Furukawa, Y.,
  Umemoto, K., Li, Y., Nakayama, H.: Synthesizing diverse lung nodules wherever
  massively: 3d multi-conditional gan-based ct image augmentation for object
  detection. In: 2019 International Conference on 3D Vision (3DV). pp.
  729--737. IEEE (2019)

\bibitem{henaff2020data}
Henaff, O.: Data-efficient image recognition with contrastive predictive
  coding. In: International Conference on Machine Learning. pp. 4182--4192.
  PMLR (2020)

\bibitem{hendrycks2019using}
Hendrycks, D., Mazeika, M., Kadavath, S., Song, D.: Using self-supervised
  learning can improve model robustness and uncertainty. arXiv preprint
  arXiv:1906.12340  (2019)

\bibitem{johnsonmimic}
Johnson, A., Lungren, M., Peng, Y., Lu, Z., Mark, R., Berkowitz, S., Horng, S.:
  Mimic-cxr-jpg-chest radiographs with structured labels

\bibitem{li2021cutpaste}
Li, C.L., Sohn, K., Yoon, J., Pfister, T.: Cutpaste: Self-supervised learning
  for anomaly detection and localization. In: Proceedings of the IEEE/CVF
  Conference on Computer Vision and Pattern Recognition. pp. 9664--9674 (2021)

\bibitem{li2020out}
Li, X., Lu, Y., Desrosiers, C., Liu, X.: Out-of-distribution detection for skin
  lesion images with deep isolation forest. In: International Workshop on
  Machine Learning in Medical Imaging. pp. 91--100. Springer (2020)

\bibitem{liang2017enhancing}
Liang, S., Li, Y., Srikant, R.: Enhancing the reliability of
  out-of-distribution image detection in neural networks. arXiv preprint
  arXiv:1706.02690  (2017)

\bibitem{mohseni2020self}
Mohseni, S., Pitale, M., Yadawa, J., Wang, Z.: Self-supervised learning for
  generalizable out-of-distribution detection. In: Proceedings of the AAAI
  Conference on Artificial Intelligence. vol.~34, pp. 5216--5223 (2020)

\bibitem{nakao2022anomaly}
Nakao, T., Hanaoka, S., Nomura, Y., Hayashi, N., Abe, O.: Anomaly detection in
  chest 18f-fdg pet/ct by bayesian deep learning. Japanese Journal of Radiology
  pp. 1--10 (2022)

\bibitem{pawlowski2018unsupervised}
Pawlowski, N., Lee, M.C., Rajchl, M., McDonagh, S., Ferrante, E., Kamnitsas,
  K., Cooke, S., Stevenson, S., Khetani, A., Newman, T., et~al.: Unsupervised
  lesion detection in brain ct using bayesian convolutional autoencoders
  (2018)

\bibitem{schlegl2019f}
Schlegl, T., Seeb{\"o}ck, P., Waldstein, S.M., Langs, G., Schmidt-Erfurth, U.:
  f-anogan: Fast unsupervised anomaly detection with generative adversarial
  networks. Medical image analysis  \textbf{54},  30--44 (2019)

\bibitem{schluter2021self}
Schl{\"u}ter, H.M., Tan, J., Hou, B., Kainz, B.: Self-supervised
  out-of-distribution detection and localization with natural synthetic
  anomalies (nsa). arXiv preprint arXiv:2109.15222  (2021)

\bibitem{seebock2019exploiting}
Seeb{\"o}ck, P., Orlando, J.I., Schlegl, T., Waldstein, S.M., Bogunovi{\'c},
  H., Klimscha, S., Langs, G., Schmidt-Erfurth, U.: Exploiting epistemic
  uncertainty of anatomy segmentation for anomaly detection in retinal oct.
  IEEE transactions on medical imaging  \textbf{39}(1),  87--98 (2019)

\bibitem{shiraishi2000development}
Shiraishi, J., Katsuragawa, S., Ikezoe, J., Matsumoto, T., Kobayashi, T.,
  Komatsu, K.i., Matsui, M., Fujita, H., Kodera, Y., Doi, K.: Development of a
  digital image database for chest radiographs with and without a lung nodule:
  receiver operating characteristic analysis of radiologists' detection of
  pulmonary nodules. American Journal of Roentgenology  \textbf{174}(1),
  71--74 (2000)

\bibitem{tan2020detecting}
Tan, J., Hou, B., Batten, J., Qiu, H., Kainz, B.: Detecting outliers with
  foreign patch interpolation. arXiv preprint arXiv:2011.04197  (2020)

\bibitem{tan2021detecting}
Tan, J., Hou, B., Day, T., Simpson, J., Rueckert, D., Kainz, B.: Detecting
  outliers with poisson image interpolation. In: Intl. Conf. Medical Image
  Computing and Computer-Assisted Intervention. pp. 581--591. Springer (2021)

\bibitem{tschuchnig2022anomaly}
Tschuchnig, M.E., Gadermayr, M.: Anomaly detection in medical imaging-a mini
  review. Data Science--Analytics and Applications pp. 33--38 (2022)

\bibitem{venkatakrishnan2020self}
Venkatakrishnan, A.R., Kim, S.T., Eisawy, R., Pfister, F., Navab, N.:
  Self-supervised out-of-distribution detection in brain ct scans. arXiv
  preprint arXiv:2011.05428  (2020)

\bibitem{wolleb2022diffusion}
Wolleb, J., Bieder, F., Sandk{\"u}hler, R., Cattin, P.C.: Diffusion models for
  medical anomaly detection. arXiv preprint arXiv:2203.04306  (2022)

\bibitem{yan2018deeplesion}
Yan, K., Wang, X., Lu, L., Summers, R.M.: Deeplesion: automated mining of
  large-scale lesion annotations and universal lesion detection with deep
  learning. Journal of medical imaging  \textbf{5}(3),  036501 (2018)

\bibitem{zenati2018efficient}
Zenati, H., Foo, C.S., Lecouat, B., Manek, G., Chandrasekhar, V.R.: Efficient
  gan-based anomaly detection. arXiv preprint arXiv:1802.06222  (2018)

\bibitem{zhao2021anomaly}
Zhao, H., Li, Y., He, N., Ma, K., Fang, L., Li, H., Zheng, Y.: Anomaly
  detection for medical images using self-supervised and translation-consistent
  features. IEEE Transactions on Medical Imaging  \textbf{40}(12),  3641--3651
  (2021)

\bibitem{zhou2020unsupervised}
Zhou, L., Deng, W., Wu, X.: Unsupervised anomaly localization using vae and
  beta-vae. arXiv preprint arXiv:2005.10686  (2020)

\end{thebibliography}

\end{document}